\documentclass{article}

    \PassOptionsToPackage{numbers, compress}{natbib}

\usepackage[preprint]{neurips_2026}

\newcommand*{\ShowNotes}{}
\usepackage{color}
\usepackage{soul}
\usepackage{multirow}
\usepackage{xcolor}
\usepackage{colortbl}
\usepackage{wrapfig}

\definecolor{darkred}{rgb}{0.7,0.1,0.1}
\definecolor{darkgreen}{rgb}{0.1,0.7,0.1}
\definecolor{cyan}{rgb}{0.7,0.0,0.7}
\definecolor{dblue}{rgb}{0.2,0.2,0.8}
\definecolor{maroon}{rgb}{0.76,.13,.28}
\definecolor{burntorange}{rgb}{0.81,.33,0}
\definecolor{tealblue}{rgb}{0.212,0.459, 0.533}
\definecolor{myyellow}{rgb}{0.8627451 , 0.75294118, 0.20784314]}

\definecolor{mypink}{rgb}{0.93359375, 0.62109375, 0.83984375}

\definecolor{pp}{rgb}{0.43921569, 0.18823529, 0.62745098}
\definecolor{rr}{rgb}{0.5254902 , 0.00784314, 0.12941176}
\definecolor{bb}{rgb}{0.09019608, 0.23529412, 0.37647059}
\definecolor{yy}{rgb}{0.49803922, 0.3372549 , 0.0}
\definecolor{gg}{rgb}{0.02352941, 0.3372549 , 0.17647059}
\definecolor{mybrown}{rgb}{0.87058824, 0.56078431, 0.01960784}
\definecolor{myblue}{rgb}{0.3372549 , 0.70588235, 0.91372549}
\definecolor{mypurple}{rgb}{0.8, 0.47058824, 0.7372549 }
\definecolor{myorange}{rgb}{0.835, 0.368, 0}
\definecolor{mygreen}{rgb}{0.00784314, 0.61960784, 0.45098039}
\definecolor{mygt}{rgb}{0.0078125 , 0.57421875, 0.40625}
\definecolor{mysp}{rgb}{0.84765625, 0.515625  , 0.0234375}
\definecolor{mycitecolor}{rgb}{0,0.08,0.45}
\definecolor{mygr}{rgb}{0.12,0.6,0.12}
\definecolor{myoo}{rgb}{0.992,0.9176,0.9019}

\definecolor{myrr}{HTML}{AE031A}
\definecolor{mybb}{HTML}{0155B3}

\definecolor{myred}{rgb}{1, 0, 0}

\definecolor{lightgray}{gray}{0.9}

\definecolor{trainable}{rgb}{1.0, 0.93, 0.78} 
\newcommand{\colorours}{\rowcolor{trainable}}

\ifdefined\ShowNotes
  \newcommand{\colornote}[3]{{\color{#1}\bf{#2: #3}\normalfont}}
\else
  \newcommand{\colornote}[3]{}
\fi

\usepackage{amsmath,amsfonts,bm}

\def\1{\bm{1}}

\def\sp{space}

\def\va{{\bm{a}}}

\def\ve{{\bm{e}}}

\def\vh{{\bm{h}}}

\def\vo{{\bm{o}}}

\def\vq{{\bm{q}}}

\def\vv{{\bm{v}}}

\def\vx{{\bm{x}}}

\def\vz{{\bm{z}}}

\def\mA{{\bm{A}}}

\def\mD{{\bm{D}}}

\def\mF{{\bm{F}}}

\def\mH{{\bm{H}}}
\def\mI{{\bm{I}}}

\def\mM{{\bm{M}}}

\def\mP{{\bm{P}}}
\def\mQ{{\bm{Q}}}

\def\mX{{\bm{X}}}
\def\mY{{\bm{Y}}}
\def\mZ{{\bm{Z}}}

\DeclareMathAlphabet{\mathsfit}{\encodingdefault}{\sfdefault}{m}{sl}
\SetMathAlphabet{\mathsfit}{bold}{\encodingdefault}{\sfdefault}{bx}{n}

\def\gC{{\mathcal{C}}}

\def\gH{{\mathcal{H}}}

\def\gO{{\mathcal{O}}}

\def\gR{{\mathcal{R}}}

\def\sR{{\mathbb{R}}}

\usepackage{symbols}
\usepackage{pifont}
\usepackage{bbm}
\usepackage{float}
\usepackage{caption}
\usepackage{graphicx}
\usepackage{enumitem}
\usepackage{amssymb}
\usepackage{wrapfig}

\newcommand{\ccmark}{{\color{green!60!black}\ding{51}}}  %
\newcommand{\cxmark}{{\color{red}\ding{55}}}             %

\newcommand{\myparagraph}[1]{\vspace*{0pt}{\bf\noindent #1}}

\newcommand\ModelName{\texttt{Div3DAffordNet}} %
\newcommand\DataName{\texttt{Div3DAfford}} %

\usepackage[utf8]{inputenc} %
\usepackage[T1]{fontenc}    %

\usepackage{url}            %
\usepackage{booktabs}       %
\usepackage{amsfonts}       %
\usepackage{nicefrac}       %
\usepackage{microtype}      %
\usepackage{xcolor}         %

\definecolor{cvprblue}{rgb}{0.21,0.49,0.74}
\definecolor{lightcarminepink}{rgb}{0.9, 0.4, 0.38}
\usepackage[pagebackref,breaklinks,colorlinks,citecolor=cvprblue,linkcolor=lightcarminepink]{hyperref}

\title{Div3DAfford: Diverse 3D affordance grounding}

\author{%
  Haomeng Zhang\quad  Raymond A. Yeh\\
  Department of Computer Science, Purdue University\\
  \texttt{\{zhan5050, rayyeh\}@purdue.edu} \\
}

\begin{document}

\maketitle
\begin{figure}[h!]
\vspace{-0.5cm}
\centering
\includegraphics[width=0.85\linewidth]{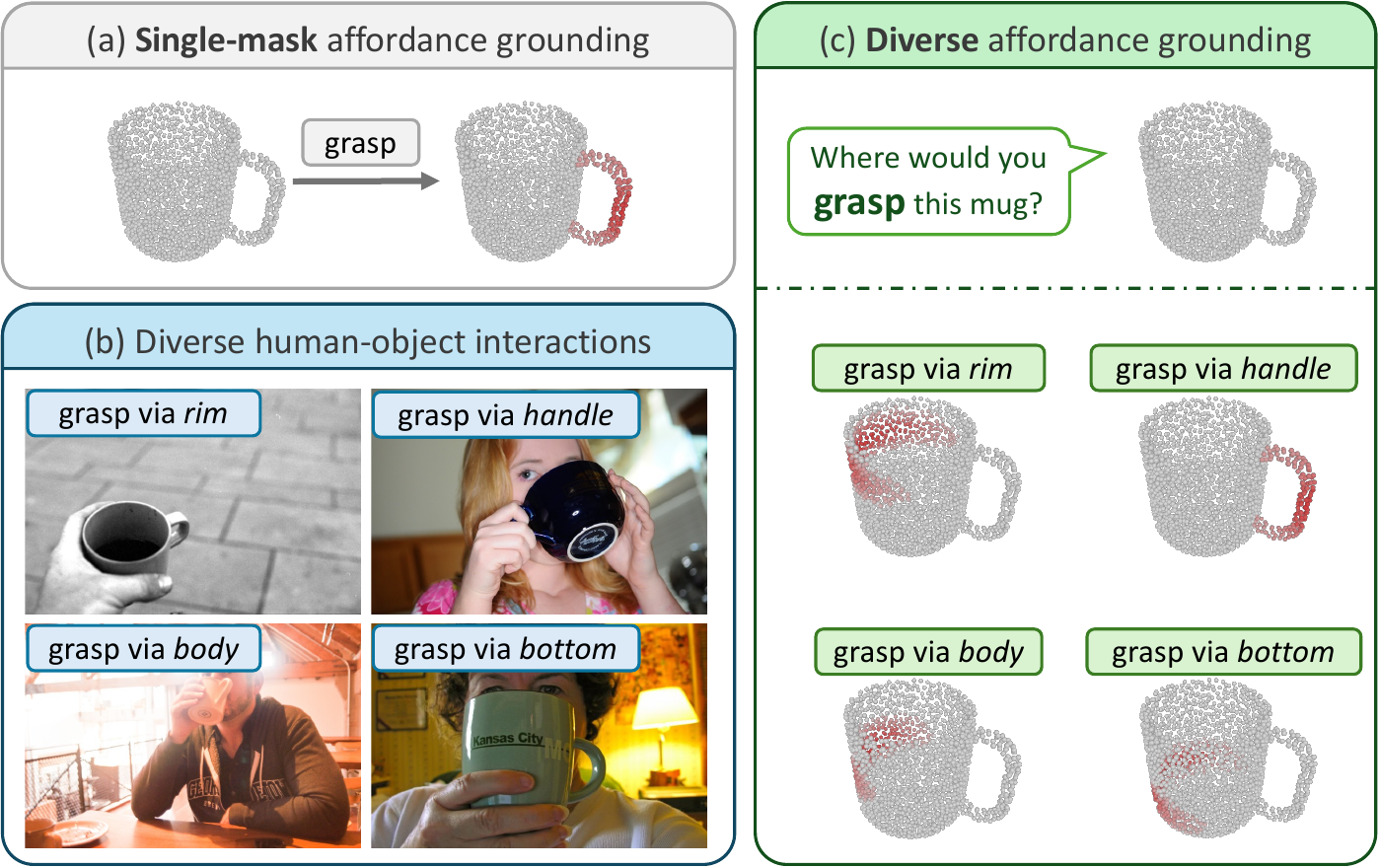}
\caption{{\bf Overview.}
\textbf{(a)} Existing 3D affordance grounding datasets and methods~\cite{deng20213d, li2024laso, chu20253d, lu2025} focus on a \emph{single-mask} setting: given an object and a query of target affordance, they predict a \textit{single} interaction region.
\textbf{(b)} However, examples from the 2D affordance dataset AGD20K~\citep{luo2022learning} illustrate that the same affordance, \eg, ``grasping a mug'', can correspond to \textit{multiple} valid regions. %
\textbf{(c)} We propose to study \emph{diverse} 3D affordance grounding that aims to predict multiple 3D affordance regions for the same object and affordance query.
}
\label{fig:teaser}
\vspace{-.1cm}
\end{figure}

\begin{abstract}
3D affordance grounding aims to localize 3D regions specified by an interaction query.
Existing methods predict a single affordance mask, overlooking the inherent diversity of human-object interactions. We introduce diverse 3D affordance grounding, a new task that requires predicting {\bf multiple plausible regions} for a given object and affordance query. We then propose \ModelName{}, a transformer-based set prediction model capable of generating diverse affordance regions. To address the scarcity of diverse 3D affordance data, we develop an automated pipeline leveraging 2D generative models and 3D reconstruction techniques to synthesize 3D affordances, followed by lightweight human verification (reducing the annotation time by 72$\times$), yielding the \DataName{} benchmark. Extensive experiments show \ModelName{} effectively captures multi-mask affordance regions. Our method achieves the SOTA affordance grounding performance in the multi-mask setting, reducing GED-SIM from 0.745 to 0.492, while maintaining competitive single-mask performance on the standard benchmark of 19.0 mIoU.
\end{abstract}

\section{Introduction}

Affordance understanding studies what actions are possible in the physical world~\citep{gibson2014ecological, visualafford}.
Rather than describing objects only by category or appearance, 3D affordance grounding captures their functional structures, \eg, a mug handle affords grasping, and a chair seat affords sitting. Such knowledge bridges perception and action by associating 3D object geometry with physically meaningful interaction regions and would enable embodied agents to interact with objects in 3D space.
This capability is crucial for various real-world applications, such as robot manipulation, human-object interaction, and virtual reality~\citep{ahn2022can, bahl2023affordances, geng2023rlafford, cheng2013affordances, li2025learning}.

Despite recent progress, existing 3D affordance grounding methods are formulated as a \emph{single-mask} task~\citep{li2024laso, lu2025, chu20253d} (see~\figref{fig:teaser}(a)): given an object-affordance pair, they predict a single affordance region. However, human-object interactions are naturally \textbf{diverse}. As illustrated in~\figref{fig:teaser}(b), the same object with the same affordance may correspond to multiple plausible regions. A mug, for example, can be grasped not only via its handle, but also via other parts such as its rim or body. This limitation arises because 3D affordance datasets~\citep{luo2022learning, zhu2025grounding, li2024laso, chu20253d} provide only a single ground-truth mask per object-affordance pair. Critically, point-level annotation on large-scale 3D assets is time-consuming, and extending such datasets through human labeling is labor-intensive and difficult to scale.

To address these challenges, we propose to study \emph{diverse 3D affordance grounding} (see~\figref{fig:teaser}(c)), a \emph{multi-mask} formulation that aims to capture multiple plausible interaction regions for the same object-affordance pair.
We introduce \ModelName{}, a transformer-based set prediction framework that decodes a set of mask queries into multiple 3D affordance masks.
Unlike single-mask methods that predict a single canonical region, \ModelName{} explicitly models multiple valid %
affordance predictions in 3D.
Besides the modeling aspect, we also need to address the data challenge: {\it how to obtain multi-mask 3D affordance annotations at scale?} For this, we propose an automated pipeline that uses 2D generative models to synthesize diverse human-object interactions and lifts them into 3D affordance annotations using 3D reconstruction models. These generated annotations allow us to extend existing single-mask labels into a multi-mask affordance dataset. %

We validate our \ModelName{} on the curated \DataName{} dataset.
For diverse 3D affordance grounding, \ModelName{} outperforms existing SOTA 3D affordance grounding methods, reducing GED-SIM from 0.745 to 0.492. Under the standard single-mask evaluation protocol, \ModelName{} remains competitive with existing single-mask methods, achieving 19.0 mIoU.

{\bf\noindent Our main contributions are as follows:}
\vspace{-3pt}
\begin{itemize}[topsep=0pt, leftmargin=12pt]
    \setlength{\itemsep}{0.0pt}
    \setlength{\parskip}{2.5pt}
    \item We propose to study \emph{diverse 3D affordance grounding}, a new task which better captures the inherent diversity of human-object interactions.
    To the best of our knowledge, this is the first work to explicitly consider multiple valid affordance regions for 3D objects.
    \item We develop \ModelName{}, a transformer-based set prediction framework which decodes spatially distributed mask queries into diverse 3D affordance masks.
    \item To obtain data, we introduce an automated pipeline that uses 2D generative models and 3D reconstruction models to generate scalable multi-mask 3D affordances to be checked by human annotators. Based on this pipeline, we construct \DataName{}, a benchmark that extends the existing single-mask 3D affordance dataset with multi-mask annotations.
\end{itemize}

\section{Related Work}
\label{sec:rel}

\myparagraph{3D affordance grounding}
aims to localize the interactive regions of 3D objects.
Early methods learn affordance directly from 3D point clouds~\citep{mo2022o2o, xu2022partafford, tabib2024lgafford}
or from 2D human-object interaction observations~\citep{luo2022learning, shao2025great}.
3D HOI works~\citep{cseke2025pico, yang2024lemon, coma2024, kim2025david, yang2024egochoir} model 3D human-object relations, contacts, and affordance priors from images or videos.
Recent methods further incorporate VLMs and LLMs for open-vocabulary 3D affordance grounding~\citep{huang2023voxposer, li2024laso, nguyen2023open, li2025intermediate, chu20253d, yu2025seqafford, van2024open, zhu2025grounding, wei20253daffordsplat, lu2025}.
Several datasets have also been introduced for different affordance grounding settings, including 2D HOI-based affordance learning~\citep{luo2022learning, shao2025great, zhu2025grounding}, open-vocabulary grounding~\citep{li2024laso, chu20253d}, and sequential affordance grounding~\citep{yu2025seqafford}.
However, the 3D affordance annotations in these datasets are still largely inherited from 3DAffordanceNet~\citep{deng20213d}, which provides a single annotated interactive region for each object-affordance pair.
As a result, existing methods typically predict one affordance region per object and do not explicitly model the inherent diversity of affordance.
In contrast, we study diverse 3D affordance grounding, where multiple plausible interaction regions may exist for the same object and affordance query.

\myparagraph{Predicting multiple outputs.}
Many vision tasks require predicting multiple plausible outputs from a single input.
One common paradigm uses generative models, such as GANs~\citep{goodfellow2014generative}, CVAEs~\cite{sohn2015learning}, flow models~\citep{lipmanflow}, and diffusion models~\citep{ho2020denoising} to capture output uncertainty through latent sampling.
Such methods have been widely adopted in trajectory prediction~\citep{gupta2018social, jiang2023motiondiffuser, salzmann2020trajectron++, shi2022motion}, human motion forecasting~\citep{aliakbarian2020stochastic, xu2022diverse, wang2024harmonizing}, stochastic video prediction~\citep{babaeizadeh2018stochastic, kumar2020videoflow}, and stochastic segmentation~\citep{kohl2018probabilistic, ribeiro2025flow}.
While expressive, these methods represent different modes implicitly in the latent space and require repeated sampling to obtain diverse predictions.
Another paradigm uses set prediction models to directly output multiple predictions. %
This formulation has been successful in object detection~\citep{carion2020end, misra2021end, zhang2024multi}, instance segmentation~\citep{cheng2022masked, cheng2021per}, object tracking~\citep{meinhardt2022trackformer, zeng2022motr}, and visual grounding~\citep{zhang2023multi3drefer, zhang2024multi}.
Our method follows this set prediction philosophy for diverse 3D affordance grounding.
Unlike stochastic generative alternatives, our query-based formulation explicitly represents multiple affordance modes without relying on latent sampling.

\myparagraph{Synthetic data from pre-trained generative models.}
Pre-trained generative models have recently been used to create synthetic supervision across vision tasks, including segmentation, detection, and amodal completion~\citep{ye2024seggen, chen2024geodiffusion, zhu2024odgen, ozguroglu2024pix2gestalt}.
Beyond these 2D settings, several works have synthesized human-object interactions for 3D reasoning.
HOI-Diffusion~\citep{zhang2024hoidiffusion} generates 3D hand-object interaction data for downstream perception, but assumes a given 3D object mesh and focuses on interaction synthesis rather than object-centric affordance.
ComA~\citep{coma2024} and DAViD~\citep{kim2025david} are more related, as both synthesize HOI observations from 3D objects to learn object-centric affordance priors.
However, these methods require clean renderable 3D meshes and perform multiview image or video synthesis for each object, making generation computationally expensive at the object level.
In contrast, our pipeline generates multi-mask 3D affordance annotations directly from 2D images.
It does not require a curated canonical mesh or expensive per-object multiview synthesis, making it more suitable for scalable instance-level annotation.

\section{Preliminaries}
\label{sec:prelim}

We briefly review the 3D affordance grounding task and the representative baseline, PointRefer~\citep{li2024laso}.

\myparagraph{Task formulation.}
Existing 3D affordance grounding is formulated as a \emph{single-mask} prediction problem~\cite{li2024laso, lu2025, li2025intermediate}.
Given a point cloud $\mP \in \sR^{N \times 3}$ along with a natural-language affordance query $A$, %
\eg, `Where would you grasp this mug?'. %
The goal is to predict a point-level affordance mask $\mY \in \sR^{N}$ that localizes the corresponding actionable region:
\bea
\label{eq:uni_afford}
\mY = F_{\text{single}}(\mP, A)
\eea
Each affordance query $A$ is associated with an affordance label $\va$ (\eg, `grasp'), which specifies the intended interaction.
This formulation implicitly assumes that each $(\mP, A)$ pair corresponds to a single affordance region.
However in practice, the same object may contain multiple spatially distinct yet equally valid regions for the queried interaction.

\myparagraph{PointRefer}~\citep{li2024laso} is a representative single-mask 3D affordance grounding model.
It first extracts hierarchical point features from the input point cloud using a PointNet++ backbone~\citep{qi2017pointnet++}.
In parallel, it encodes the text query using a pre-trained RoBERTa~\citep{liu2019roberta} text encoder.
The text features are injected into the point features through repeated language-conditioned adaptive fusion modules.
The resulting point features are fused with the text tokens through a transformer decoder, producing a point-wise feature map $\mF \in \sR^{N \times C}$.
Let $\mH = [\vh_1,\dots,\vh_L] \in \sR^{L \times C}$ denote the decoded text features, where $L$ is the token length.
PointRefer predicts a single affordance mask by computing token-to-point similarity and averaging across valid text tokens:
\begin{equation}
\mY = \sigma\left(
\frac{1}{\sum_{l=1}^{L} m_l}
\sum_{l=1}^{L} m_l\vh_l\mF^\top
\right),
\end{equation}
where $m_l \in \{0,1\}$ denotes the attention mask and $\sigma(\cdot)$ is the sigmoid function.
While effective for single-mask prediction, this formulation produces only one mask and therefore cannot explicitly represent multiple valid affordance regions for the same object.

\section{Approach}
\label{sec:approach}

Different from single-mask 3D affordance grounding in~\equref{eq:uni_afford}, which predicts a single affordance mask, we formulate diverse 3D affordance grounding as a \emph{multi-mask} (set) prediction task:
\bea
\label{eq:multi_afford}
\{\mY_{s}\}_{s=1}^{S} = F_{\text{multi}}(\mP, A),
\eea
where each $\mY_{s} \in \sR^{N}$ represents a valid mask for the 3D object and affordance query.
The number of valid masks $S$ varies across different object-query pairs. This formulation allows the model to predict multiple spatially distinct yet plausible interaction regions for the same input.

To build an effective model for this multi-mask task, we need to answer the following questions:
\vspace{-3pt}
\begin{itemize}[topsep=0pt, leftmargin=12pt]
  \item {\bf How to design a model architecture that can predict a variable-size set of affordance masks}, and encourage different predictions to capture different interaction modes? We introduce \ModelName{}, a transformer-based multi-mask prediction framework that maintains a set of mask queries to decode multiple affordance masks in parallel, and is trained with a set prediction objective to learn both \emph{what} masks to predict and \emph{how many} valid modes to output (\secref{sec:model}).
  \item {\bf How to collect large-scale multi-mask 3D affordance annotations} for training such a model, given that existing annotations are single-mask? We leverage 2D generative models and 3D reconstruction techniques to produce multi-mask 3D affordance annotations from 2D images, enabling the construction of the \DataName{} dataset (\secref{sec:data_gen}).
\end{itemize}

\begin{figure}[t]
\centering
\includegraphics[width=\linewidth]{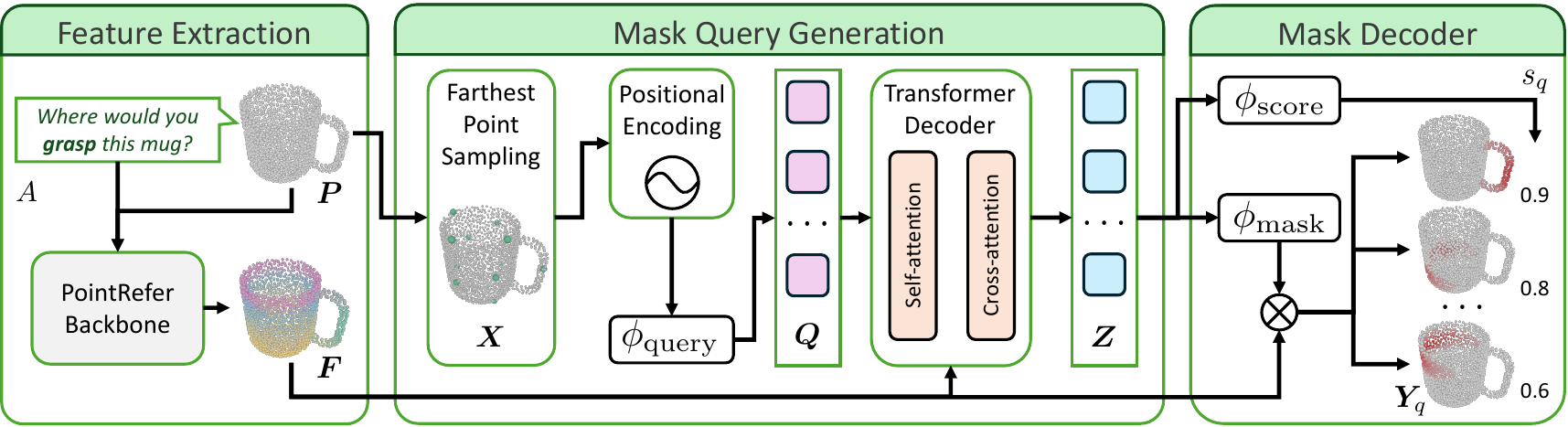}
\caption{
\textbf{\ModelName{} pipeline (\secref{sec:model}).}
Given a point cloud $\mP$ and an affordance query $A$, a PointRefer backbone extracts language-conditioned point-wise features $\mF$.
Mask queries $\mQ$ are initialized from FPS seeds $\mX$ and refined with a transformer decoder.
Each refined mask query $\vz_q$ predicts a probability score $s_q$ and a mask embedding, which is correlated with the shared point-wise features to produce a candidate affordance mask $\mY_q$.
}
\label{fig:model_pipeline}
\vspace{-0.3cm}
\end{figure}

\subsection{\ModelName{}}
\label{sec:model}
Inspired by DETR from object detection~\citep{carion2020end,misra2021end}, we equip PointRefer~\citep{li2024laso} with a query-based multi-mask decoder to support the variable-size set prediction.
We name the overall model \ModelName{} and provide a visual overview in~\figref{fig:model_pipeline}.

\myparagraph{Multi-mask decoder.} To prepare the input for this decoder, we follow PointRefer~\cite{li2024laso} to extract point-wise features $\mF \in \sR^{N \times C}$ from an input point cloud $\mP$ and an affordance query $A$ using PointNet++ encoding and language-guided adaptive fusion. Next, the multiple plausible affordances in \ModelName{} are represented by maintaining a set of $Q$ mask query tokens.
Each query corresponds to one candidate affordance and is decoded into an affordance mask along with a score indicating whether the affordance is valid. 

That is, the decoder outputs a set of candidate affordance masks $\{\mY_q\}_{q=1}^{Q}$ together with scores $\{s_q\}_{q=1}^{Q}$. %
At prediction time, we retain the subset of masks whose score exceeds a threshold $\tau_s$:
\bea
\{\mY_q \mid s_q \ge \tau_s\}. %
\eea

In more detail, each mask $\mY_q$ and score $s_q$ are computed from a \emph{refined} mask query $\vz_q$, a token representation that encodes a candidate affordance after attending to the point features (detailed later in \myparagraph{Mask query generation}).
Each $\vz_q$ predicts a mask embedding $\ve_q$ and a score $s_q$, \ie,
\bea
\ve_q = \phi_{\text{mask}}(\vz_q),
\qquad
s_q = \sigma(\phi_{\text{score}}(\vz_q)),
\eea
where $\sigma$ denotes the sigmoid function and $\phi_{\text{mask/score}}$ are lightweight MLPs.
The $q$-th affordance mask 
\bea
\mY_q
=
\sigma(\ve_q\mF^\top)
\eea
is then obtained by correlating the mask embedding with the point-wise feature map.

\myparagraph{Mask query generation.}
To generate diverse affordance candidates, the set of mask queries is initialized from ``3D seeds'' corresponding to different spatial regions. We sample $Q$ query seeds from the input point cloud using farthest point sampling~\citep{qi2017pointnet++}, which encourages broad coverage of the object surface.
Let $\mX = \{\vx_q\}_{q=1}^{Q}$ denote the 3D seed coordinates.
Following prior work~\citep{misra2021end}, each seed coordinate $\vx_q$ is encoded with Fourier positional embedding~\citep{tancik2020fourier} and projected by an MLP to obtain the initial mask query $\vq_q$.
This initialization gives each mask query a distinct spatial anchor, which helps the model to focus on different affordance regions. %

Next, we refine $\vq_q$ using a transformer decoder~\cite{vaswani2017attention} to be conditioned on the language-conditioned point features $\mF$, \ie,
\bea
\mZ
=
\operatorname{Transformer-Decoder}\!\left(\mQ, \mF\right),
\eea
where $\mQ = \{\vq_q\}_{q=1}^{Q}$ are the initial mask queries and $\mZ = \{\vz_q\}_{q=1}^{Q}$ are the refined mask queries.
Each decoder layer contains self-attention among queries and cross-attention from mask queries to point features.
Each mask query selectively aggregates features from different regions of an object, allowing the model to capture multiple plausible affordance modes from the same input.

\myparagraph{Training objective.}
In the training set, each sample has a set of ground-truth affordance masks $\mathcal{Y}=\{\mY^*_k\}_{k=1}^{K}$, where $ 1 \leq K \le Q$. As sets are unordered, we need to decide which ground-truth mask $\mY^*_k$ should be used to supervise the $q$--th predicted mask $\mY_q$ from our model.

For this, we introduce a DETR-style bipartite matching objective. First, 
we create a one-to-one matching between the $K$ ground-truth masks and the $Q$ predicted masks. Let $\pi$ denote an injective assignment from ground-truth indices to mask query indices.
We solve for the optimal assignment by minimizing a matching cost $\gC$:
\bea
\pi^{\star}
=
\arg\min_{\pi}
\sum_{k=1}^{K}
\mathcal{C}\bigl(\mY^*_k,\mY_{\pi(k)},s_{\pi(k)}\bigr).
\eea
The matching cost considers both mask reconstruction quality and predicted score:
\bea
\label{eq:matching_cost}
\mathcal{C}(\mY^{*}_k,\mY_q, s_q) = 
\alpha_{\text{bce}}\ell_{\text{bce}}(\mY^{*}_k,\mY_q)
+
\alpha_{\text{dice}}\ell_{\text{dice}}(\mY^{*}_k,\mY_q)
-
\alpha_{\text{score}}\log(s_q),
\eea
where $\alpha_{\text{bce}/\text{dice}/\text{score}} \in \sR^+$ and $\ell_{\text{bce}/\text{dice}}$ correspond to the binary cross entropy loss, and the dice loss~\cite{milletari2016v}.

With the optimal matching $\pi^{\star}$, the loss for one sample is:
\bea
\label{eq:set_loss}
\begin{aligned}
\mathcal{L}_{\text{set}}
=
&
\sum_{k=1}^{K}
\Big[
\lambda_{\text{score}}\big(-\log s_{\pi^{\star}(k)}\big)
+
\lambda_{\text{bce}}\ell_{\text{bce}}\big(\mY^{*}_k,\mY_{\pi^{\star}(k)}\big)
+
\lambda_{\text{dice}}\ell_{\text{dice}}\big(\mY^{*}_k,\mY_{\pi^{\star}(k)}\big)
\Big]
\\
&+
\lambda_{\text{nomask}}
\sum_{q \notin \operatorname{Im}(\pi^{\star})}
-\log(1-s_q),
\end{aligned}
\eea
where $\operatorname{Im}(\pi^{\star})$ denotes the set of matched query indices and $\lambda$s are hyperparameters balancing the individual loss terms. Intuitively, the matched queries are encouraged to be more confident and accurate, while the unmatched queries are suppressed to have a lower score. This objective enables the model to learn both \emph{what} masks to predict and \emph{how many} valid hypotheses to output. See Appendix~\secref{sec:model_details} for more details.

\begin{figure}[t]
\centering
\includegraphics[width=\linewidth]{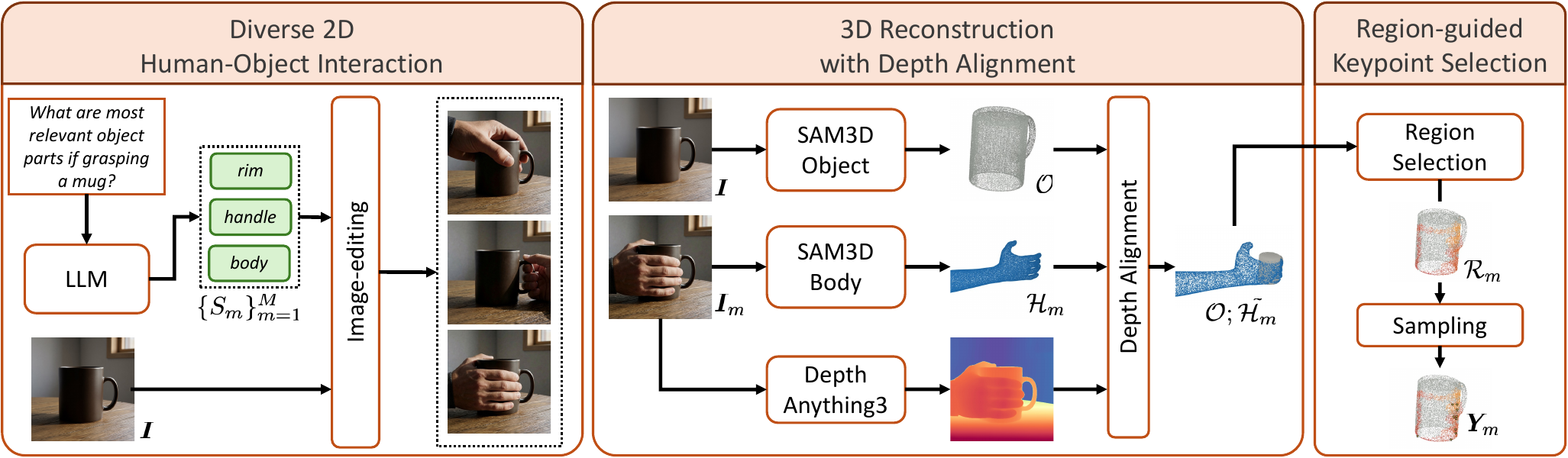}
\caption{
\textbf{Annotation generation pipeline (\secref{sec:data_gen}).}
Given an object image $\mI$, we generate diverse 2D HOI images using LLM-proposed interaction subparts $\{S_{m}\}_{m=1}^{M}$ and an image-editing model.
From the image $\mI$ and each HOI image $\mI_m$, we reconstruct the 3D object $\gO$ and human/hand $\tilde{\gH_m}$ with depth alignment, then extract keypoints and propagate them into soft 3D affordance labels $\mY_m$.
}
\label{fig:data_gen_pipeline}
\vspace{-.5cm}
\end{figure}

\subsection{Obtaining Multi-mask Affordance Data}
\label{sec:data_gen}
Existing 3D affordance annotations require substantial human efforts to collect and remain largely single-mask~\cite{deng20213d, luo2022learning, shao2025great, zhu2025grounding, chu20253d, yu2025seqafford}.
3DAffordanceNet~\cite{deng20213d} reports hiring 42 professional annotators, with an average annotation time of 2 minutes per shape and each shape annotated by 3 annotators~\citep{deng20213d}.
To scale multi-mask 3D affordance annotation, we introduce a data curation pipeline (\figref{fig:data_gen_pipeline}) that constructs annotations for our proposed task. The pipeline is based on recent advancements in 2D generative models and 3D reconstruction techniques to produce high-quality multi-mask 3D affordance annotations from 2D images.

Importantly, the pipeline {\bf shifts the role of human annotators from point-level annotation to lightweight inspection and filtering.}
We manually inspect the generated annotations and filter out unreasonable samples, such as failed reconstructions and severe human-object misalignment.
This human-in-the-loop validation takes about \textit{five} seconds per sample with a single annotator, corresponding to roughly a 72$\times$ reduction in human annotation time compared with the reported 2-minute annotation time by~\citet{deng20213d}.

\myparagraph{Diverse 2D human-object interaction.}
Given a 2D image with its semantic class and target affordance label, we first generate a set of plausible 2D interaction patterns.
Specifically, we prompt an LLM (Qwen3~\cite{wu2025qwen}) to propose a set of plausible object subparts $\{S_{m}\}_{m=1}^{M}$ that may support the target interaction.
To ``grasp a mug'', the LLM may suggest subparts such as ``rim'', ``handle'', and ``body''.
For each suggested subpart $S_m$, we then use a 2D image editing model (Qwen3-Image-Edit-2511~\cite{wu2025qwen}) to synthesize an HOI image $\mI_m$ in which a human interacts with the object at the corresponding region.
This produces a set of HOI images $\{\mI_m\}_{m=1}^{M}$, capturing $M$ candidate interaction patterns for the same object.

\myparagraph{3D reconstruction with depth alignment.}
From the original image $\mI$ and each synthesized HOI image $\mI_{m}$, we reconstruct the object and human using SAM3D-Object~\citep{sam3dobject} and SAM3D-Body~\citep{yang2026sam}, obtaining an object point cloud $\gO $ and a human point cloud $\gH_m$.
Since the two point clouds are reconstructed independently, their relative depths may be inconsistent.
We use monocular depth from DepthAnything3~\cite{lin2025depth} to align them in a shared camera coordinate system.
Next, we fit affine mappings between reconstructed depth and monocular depth for the object and human, and transform the human point cloud along its camera rays into the object depth scale.
This yields an aligned human point cloud $\tilde{\gH_m}$ that is spatially consistent with $\gO$.

\myparagraph{Region-guided keypoint selection.}
Given the aligned object-human point clouds $(\gO,\tilde{\gH_m})$, we estimate the interaction region $\gR_m$ by selecting object points that are spatially close to the human.
We sample $P$ keypoints from $\gR_m$ and propagate them following prior 3D affordance annotation protocols~\cite{deng20213d}, producing a soft affordance mask $\mY_m$ for each HOI image $\mI_m$.
Aggregating all masks gives the multi-mask annotation set $\{\mY_m\}_{m=1}^{M}$.
Details are included in~\secref{sec:data_gen_details} in the appendix.

\begin{wraptable}[11]{r}{0.45\linewidth}
\vspace{-.45cm}
\centering
\caption{
\textbf{Ablation study of the generation pipeline on 3DAffordanceNet validation set.}
We report the normalized Chamfer Distance (CD).
Acronyms: DA.: depth alignment; 
KS.: region-guided keypoint selection.
\cxmark{} for KS. means directly using the points closest to the human point cloud as keypoints.
}
\vspace{-3pt}
    \begin{tabular}{ccc}
    \specialrule{.15em}{.05em}{.05em}
    DA. & KS. & Norm. CD ($\downarrow$) \\
    \hline
    \cxmark & \cxmark & 0.328 \\
    \ccmark & \cxmark & 0.282 \\
    \ccmark & \ccmark & \textbf{0.187} \\
    \specialrule{.15em}{.05em}{.05em}
    \end{tabular}
\vspace{-8pt}
\label{tab:ablation_data_gen}
\end{wraptable}

\vspace{3pt}
\textbf{Constructing the \DataName{} benchmark.} %
We create the \DataName{} dataset by first rendering 2D views of each point cloud from 3DAffordanceNet~\cite{deng20213d}, and then applying our annotation generation pipeline to these rendered views. 
Since our generated annotations are defined on the
SAM3D reconstruction, we align the reconstructed 3D object with the original 3DAffordanceNet shape using the Iterative Closest Point (ICP) algorithm~\cite{besl1992method}, and transfer the extracted affordance labels to the original shape via nearest-neighbor assignment.
We then augment the original single-mode annotation with our generated affordance annotations, resulting in \DataName{}.
For each object-affordance pair, we adopt the natural-language affordance query from LASO~\citep{li2024laso}.
The dataset contains 14,606 training samples and 4,452 validation samples spanning 55 object-affordance pairs.
On average, each shape is associated with 3.04 annotated masks, providing multi-mask annotation for multiple plausible interaction regions.

\begin{wrapfigure}[20]{r}{0.48\linewidth}
\vspace{-.15cm}
\centering
\includegraphics[width=\linewidth]{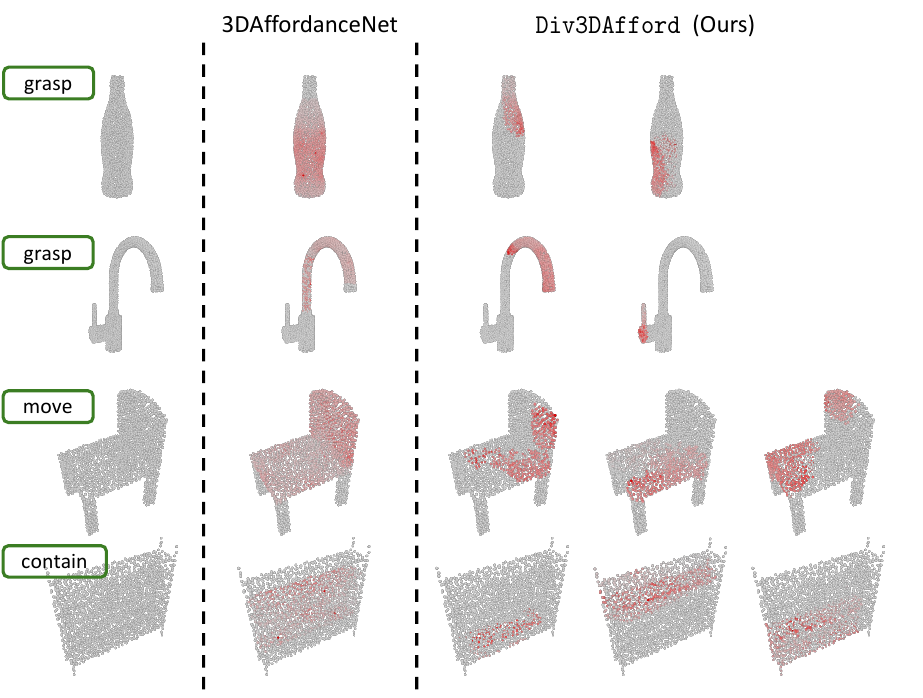}
\caption{
\textbf{\DataName{} data samples.}
We visualize representative 3D affordance annotations from \DataName{}.
Our dataset extends the original 3D AffordanceNet annotations with additional plausible interaction modes.
}
\label{fig:qual_data_gen}
\vspace{-.2cm}
\end{wrapfigure}

\vspace{3pt}
\textbf{Evaluating \DataName{} quality.}
We validate \DataName{} on the 3DAffordanceNet validation set by comparing our automatically generated affordance keypoints with the human-annotated affordance keypoints from 3DAffordanceNet. 
We measure their geometric consistency using the normalized Chamfer Distance (CD)~\cite{achlioptas2018learning}. %
As shown in~\tabref{tab:ablation_data_gen}, both depth alignment and region-guided keypoint selection improve the quality of the generated annotations, demonstrating the effectiveness of our annotation generation pipeline.
Note that the human annotations in 3DAffordanceNet reflect only one annotator-preferred interaction mode.
Therefore, our goal is not to exactly reproduce this single mode, but to generate additional valid interaction modes for the same object-affordance pair.
As shown in~\figref{fig:qual_data_gen}, our generated annotations cover the original 3DAffordanceNet mode, while also capturing additional plausible regions. %
Together with manual inspection and filtering of the generations, our pipeline effectively produces diverse and valid 3D affordance annotations.
Please refer to~\secref{sec:dataset_details} for evaluation details and additional results.

\section{Experiments}
\label{sec:experiments}
We evaluate the proposed \ModelName{} on both multi-mask and single-mask 3D affordance grounding.
Our experiments are designed to answer three questions:
{\bf (a)} whether \ModelName{} can predict multiple plausible 3D affordance masks for the same object and affordance query (\secref{sec:exp_multi});
{\bf (b)} whether \ModelName{} trained with multi-mask supervision can still capture the specific annotation mode evaluated by standard single-mask benchmarks (\secref{sec:exp_uni}); and
{\bf (c)} how the query-based set prediction design affects mode coverage and prediction quality (\secref{sec:exp_query}).

\myparagraph{Baselines.}
We compare \ModelName{} with representative 3D affordance grounding baselines, including PointRefer~\cite{li2024laso}, 3D-ADLLM~\cite{chu20253d}, and GEAL~\cite{lu2025}.
Note that we directly evaluate 3D-ADLLM's official checkpoint, as we were unable to reproduce the training.

Next, we further implement two generative multi-mask baselines using the PointRefer backbone: PointRefer-CVAE and PointRefer-LDM.
PointRefer-CVAE incorporates a conditional variational autoencoder~\cite{sohn2015learning} for diverse mask generation.
PointRefer-LDM uses a latent diffusion decoder~\cite{rombach2022high} to generate multiple affordance predictions.
See Appendix~\secref{sec:baseline_details} for details of all baselines.

\subsection{Multi-mask evaluation}
\label{sec:exp_multi}

\myparagraph{Evaluation metrics.}
We evaluate multi-mask prediction quality using Generalized Energy Distance (GED)~\cite{ribeiro2025flow}, which measures the discrepancy between the predicted mask distribution and the ground-truth mask distribution.
We use standard single-mask affordance metrics as the underlying distance functions, and report the corresponding GED scores.
Lower GED indicates that the predicted mask set better matches the ground-truth multi-mask annotation set.
Please refer to~\secref{sec:supp_eval_metrics} for details.

\myparagraph{Quantitative results.}
We show quantitative results in~\tabref{tab:multimodal_main}.
Compared with the single-mask baselines, the multi-mask generative baselines and our \ModelName{} achieve better multi-mask grounding performance.
While deterministic methods can learn to localize a single annotation mode, they do not capture multiple valid modes.
Generative variants PointRefer-CVAE and PointRefer-LDM improve diversity by sampling multiple predictions. Overall,  our \ModelName{} achieves the best performance. 

We suspect that these generative baselines are less effective because diverse 3D affordance grounding contains only a small set of discrete interaction modes. Typically, generative models~\citep{rombach2022high} are designed to capture much greater diversity, making their architectures an excessive choice for this task. In contrast, our proposed \ModelName{} explicitly models this small number of modes, resulting in a more intuitive and cleaner design.

\begin{table}[t]
\centering
\caption{
\textbf{Quantitative results of diverse 3D affordance grounding.}
We report Generalized Energy Distance (GED)~\cite{ribeiro2025flow} with modified distance metrics tailored for 3D affordance grounding.
$^{*}$ indicates direct inference with the official checkpoint.
}
\begin{tabular}{lccc}
\specialrule{.15em}{.05em}{.05em}
Method & GED-AUC ($\downarrow$) & GED-MAE ($\downarrow$) & GED-SIM($\downarrow$) \\
\hline
3D-ADLLM$^{*}$~\cite{chu20253d}    & 0.731 & 1.712  & 0.973 \\
GEAL~\cite{lu2025}           & 0.418 & 0.391 & 0.782 \\
PointRefer~\cite{li2024laso}       & 0.337 & 0.169 & 0.745 \\
\hline
PointRefer-LDM & 0.371 & 0.120 & 0.636 \\
PointRefer-CVAE & 0.415 & \textbf{0.096} & 0.602 \\
\colorours \ModelName   & \textbf{0.254} & \textbf{0.096} & \textbf{0.492} \\
\specialrule{.15em}{.05em}{.05em}
\end{tabular}
\label{tab:multimodal_main}
\vspace{-0.2cm}
\end{table}

\begin{figure}[t]
\centering
\includegraphics[width=\linewidth]{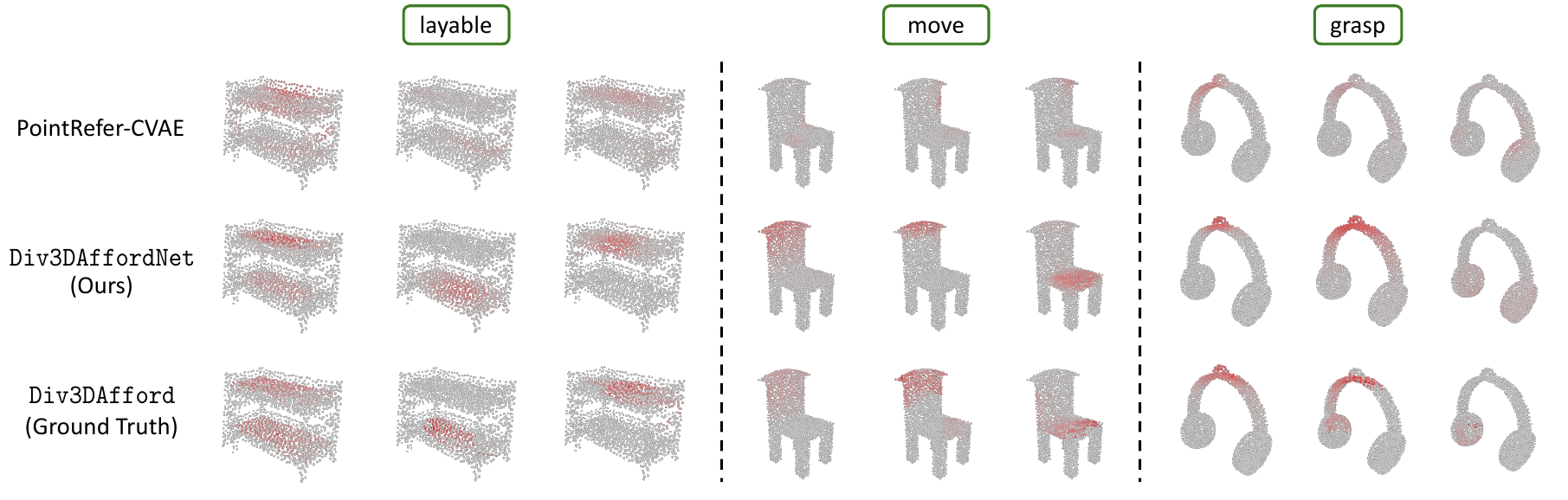}
\caption{\textbf{Qualitative comparison on diverse 3D affordance grounding.}
We compare \ModelName{} with the PointRefer-CVAE baseline on the \DataName{} validation set.
For each object-affordance query, we visualize representative predictions for different output modes.
Red indicates the predicted affordance region.
}
\label{fig:qual_multi_results}
\vspace{-0.5cm}
\end{figure}

\begin{table}[t]
\centering
\caption{
\textbf{Quantitative results of single-mask 3D affordance grounding.}
We report Area Under the Curve (AUC), Mean Intersection over Union (mIoU), Similarity (SIM), and Mean Absolute Error (MAE).
$^{*}$ indicates direct inference with the official checkpoint.
}
\begin{tabular}{lcccc}
\specialrule{.15em}{.05em}{.05em}
Method & AUC ($\uparrow$) & MAE ($\downarrow$) & mIoU ($\uparrow$) & SIM ($\uparrow$) \\
\hline
3D-ADLLM$^{*}$~\cite{chu20253d}    & 0.500 & 0.388 & 3.6 & 0.399 \\
GEAL~\cite{lu2025}           & 0.824 & 0.132 & 18.2 & 0.567 \\
PointRefer~\cite{li2024laso}       & \textbf{0.858} & 0.096 & 18.8 & 0.617\\
\colorours \ModelName   & 0.842 & \textbf{0.093} & \textbf{19.0} & \textbf{0.623} \\
\specialrule{.15em}{.05em}{.05em}
\end{tabular}
\label{tab:unimodal_main}
\end{table}

\myparagraph{Qualitative results.}
We provide qualitative comparisons on the \DataName{} validation set in~\figref{fig:qual_multi_results}.
Our \ModelName{} produces diverse and spatially coherent affordance masks that correspond to distinct plausible interaction regions.
For example, our \ModelName{} accurately identifies that an earphone can be grasped at different regions, including different parts of the headband and the ear pads.
In contrast, PointRefer-CVAE often produces unconfident and imprecise predictions.

\subsection{Single-mask evaluation}
\label{sec:exp_uni}
We further evaluate whether our \ModelName{} trained on multi-mask supervision remains compatible with standard single-mask benchmarks. We compare our \ModelName{} against single-mask baselines under the conventional single-mask 3D affordance grounding protocol.
We report results on the human-annotated affordance data from 3DAffordanceNet.

\myparagraph{Evaluation metrics.} Since \ModelName{} predicts a set of plausible masks,
we use an oracle best-mask selection based on MAE to measure whether the predicted set contains the human-annotated mode.
This evaluation measures whether \ModelName{} can accurately capture the evaluated annotation mode without being distracted by other plausible affordance masks.
Following prior work~\citep{li2024laso}, we report four standard single-mask affordance grounding metrics: Area Under the Curve (AUC), mean Intersection over Union (mIoU), Similarity (SIM), and Mean Absolute Error (MAE).
See~\secref{sec:supp_eval_metrics} for details.

\myparagraph{Quantitative results.}
We show the quantitative results in~\tabref{tab:unimodal_main}.
The single-mask baselines are trained and tested on the same human-annotated single-mask annotations, whereas \ModelName{} is trained with additional multi-mask annotations from \DataName{}.
We observe that \ModelName{} remains competitive under the standard single-mask evaluation protocol.
This suggests that %
\ModelName{} can predict multiple interaction modes while producing accurate masks for the annotation mode evaluated by the conventional single-mask benchmark.

\begin{table}[t]
\centering
\caption{
\textbf{Ablation study on the number of query seeds $Q$.}
We vary the number of query seeds and report GED-based metrics on multi-mask prediction.
}
\begin{tabular}{lccc}
\specialrule{.15em}{.05em}{.05em}
$Q$ & GED-AUC ($\downarrow$) & GED-MAE ($\downarrow$) & GED-SIM ($\downarrow$) \\
\hline
16  & 0.288 & 0.099  & 0.506 \\
\colorours 32  & \textbf{0.254} & \textbf{0.096}  & \textbf{0.492} \\
64  & 0.268 & 0.097 & 0.497 \\
128  & 0.271 & 0.100 & 0.499 \\
\specialrule{.15em}{.05em}{.05em}
\end{tabular}
\label{tab:ablation_query_num}
\end{table}

\subsection{Analysis on the effect of query seeds}
\label{sec:exp_query}

We further study the effect of the number of query seeds $Q$ in~\tabref{tab:ablation_query_num}.
Using too few queries limits the model's capacity to cover diverse affordance modes, while using too many queries can introduce redundant or low-quality candidate masks.
Empirically, $Q=32$ achieves the best performance across all GED-based metrics and provides a good balance between coverage and prediction quality.

\begin{figure}[h]
\vspace{-0.4cm}
\centering
\includegraphics[width=0.98\linewidth]{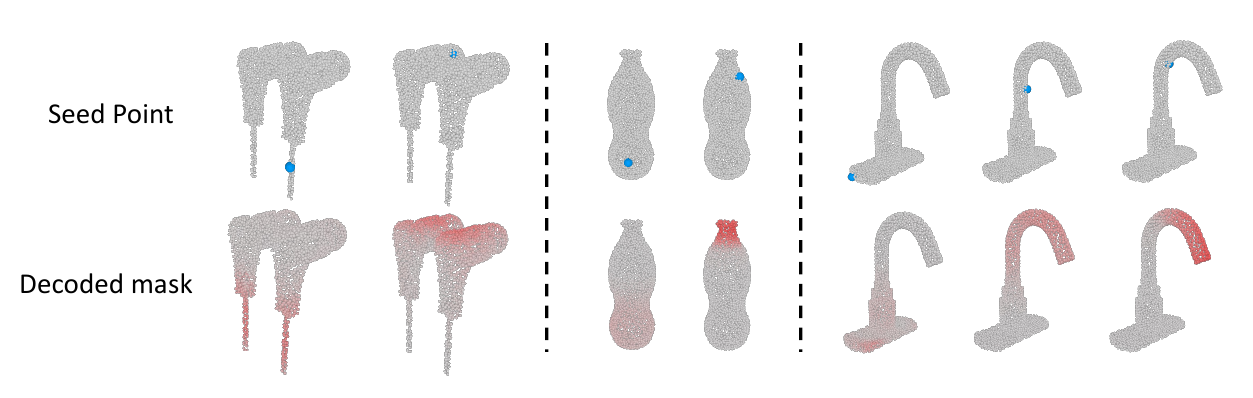}
\vspace{-0.35cm}
\caption{
\textbf{Qualitative analysis of query seeds.}
We visualize the query seeds (blue) and their decoded affordance masks (red) for representative examples.
Each query acts as a spatial anchor, and its decoded mask is localized to a nearby valid interaction region.
}
\label{fig:query_analysis}
\vspace{-0.25cm}
\end{figure}

We analyze the effect of query seeds on our \ModelName{}.
As shown in~\figref{fig:query_analysis}, each query seed acts as a spatial anchor for mask decoding.
The decoded affordance mask is typically localized to a nearby valid interaction region, indicating that different queries specialize in different plausible affordance modes.
This is desirable for diverse 3D affordance grounding, where the same object and affordance query may correspond to multiple valid interaction regions.

\section{Conclusion}
This work introduces \emph{diverse} 3D affordance grounding, which studies the inherent diversity of how humans interact with 3D objects.
We propose \ModelName{}, a query-based diverse 3D affordance grounding model that
uses spatially distributed mask queries to predict multiple plausible interaction regions for a given object and affordance prompt.
Due to the lack of multi-mask 3D affordance annotations, we present a data curation pipeline that leverages 2D generative models to generate multi-mask annotations to be reviewed by human annotators. This procedure reduces the annotation time by $72\times$, leading to the \DataName{} dataset. %
Extensive experiments demonstrate that \ModelName{} successfully captures multiple valid affordance modes while remaining highly competitive on standard single-mask evaluations.
We hope our framework and dataset will inspire future research in diverse and comprehensive 3D affordance modeling.

\clearpage

{
\small
\bibliographystyle{ieeenat_fullname}
\bibliography{main}
}

\clearpage
\section*{Appendix}
\setcounter{section}{0}
\renewcommand{\theHsection}{A\arabic{section}}
\renewcommand{\thesection}{A\arabic{section}}
\renewcommand{\thetable}{A\arabic{table}}
\setcounter{table}{0}
\setcounter{figure}{0}
\setcounter{equation}{0}
\renewcommand{\thetable}{A\arabic{table}}
\renewcommand\thefigure{A\arabic{figure}}
\renewcommand{\theHtable}{A.Tab.\arabic{table}}%
\renewcommand{\theHfigure}{A.Abb.\arabic{figure}}%
\renewcommand\theequation{A\arabic{equation}}
\renewcommand{\theHequation}{A.Abb.\arabic{equation}}%

\noindent The appendix is organized as follows:
\begin{itemize}
\item In~\secref{sec:model_details}, we provide implementation details of \ModelName{}.
\item In~\secref{sec:data_gen_details}, we provide details on the data generation process.
\item In~\secref{sec:dataset_details}, we provide additional \DataName{} details.
\item In~\secref{sec:baseline_details}, we provide implementation details of the baselines.
\item In~\secref{sec:supp_eval_metrics}, we provide details on the evaluation metrics.
\item In~\secref{sec:add_qual_results}, we provide additional qualitative results.
\item In~\secref{sec:add_ablation_studies}, we provide additional ablation studies.
\item In~\secref{sec:compute}, we provide the computational cost.
\item In~\secref{sec:limitations}, we discuss the limitations of our approach.
\end{itemize}

\section{\ModelName{} implementation details}
\label{sec:model_details}
In this section, we provide additional implementation details of our \ModelName{}.

\myparagraph{Architecture hyperparameters.}
For the PointRefer backbone~\citep{li2024laso}, we use $2048$ input points, a feature dimension of $512$, $4$ attention heads, and a RoBERTa-base text encoder~\citep{liu2019roberta}.
For mask query generation, we use $Q=32$ mask queries and $2$ transformer decoder layers.
During inference, we select masks whose query probability is above $\tau_s=0.3$ and always keep at least the highest-scoring query to avoid empty prediction sets.

\myparagraph{Training hyperparameters.}
The matching-cost weights in ~\equref{eq:matching_cost} are set to $\alpha_{\text{score}}=1.0$, $\alpha_{\text{bce}}=1.0$, and $\alpha_{\text{dice}}=1.0$.
The loss weights in~\equref{eq:set_loss} are set to $\lambda_{\text{score}}=1.0$, $\lambda_{\text{bce}}=1.0$, $\lambda_{\text{dice}}=1.0$, and $\lambda_{\text{nomask}}=0.1$.
We train the model for $50$ epochs on a single NVIDIA H100 GPU, using Adam with betas $(0.9,0.999)$.
The learning rate is decayed by a factor of $0.5$ every $10$ epochs using a step scheduler.

\section{Generation pipeline details}
\label{sec:data_gen_details}
\myparagraph{Diverse 2D interaction synthesis.}
Given an object class $c$ and an affordance query $a$, we first render the input 3D object into a 2D image from a selected viewpoint.
Since direct point-cloud renderings often contain sparse visual artifacts, we use Qwen-Image-Edit~\citep{wu2025qwen} to convert each synthetic rendering into a photorealistic object image.
Specifically, we use the following image-editing prompt:
\begin{quote}
\small
\textit{This image is a synthetic point cloud rendering of a $\{c\}$. Convert it into a photorealistic image of a real $\{c\}$. Preserve the exact object shape, proportions, and viewpoint. Replace the point cloud appearance with realistic materials, textures, and surface details. Use natural lighting, realistic shadows, and a real camera look.}
\end{quote}
This produces a clean object image $\mI$ that remains geometrically consistent with the original 3D shape.
To synthesize diverse interaction patterns, we first ask an LLM to propose a small set of affordance-relevant object parts for each object-affordance pair.
For object class $c$ and affordance $a$, we query the LLM with:
\begin{quote}
\small
\textit{Given the affordance $\{a\}$ and object category $\{c\}$, list the $\{M\}$ most relevant object parts that would be involved in this affordance. Provide your answer as a comma-separated list of parts only, without any additional explanation. For example: `handle, rim, body'.}
\end{quote}
The LLM returns $\{S_{m}\}_{m=1}^{M}$, where each $S_m$ denotes a plausible object part involved in the interaction, such as a handle, rim, or body.
Each part then defines one interaction mode.
For each $S_m$, we use Qwen-Image-Edit-2511 to edit $\mI$ into an interaction image $\mI_m$ by adding an interaction cue localized around the specified part.
Most affordances are synthesized as human-object interactions.
Depending on the affordance type, we use either a hand-level interaction prompt or a full-body interaction prompt.
For hand-centric affordances, we use:
\begin{quote}
\small
\textit{Edit this image to add a realistic human hand $\{a\}$ the $\{S_m\}$ of the $\{c\}$. Only show the hand and part of the wrist, not the full person. The hand should be in the foreground with natural contact and shadows. Do not alter the $\{c\}$'s geometry, position or components.}
\end{quote}
For body-centric affordances, we use:
\begin{quote}
\small
\textit{Edit this image to add a realistic full human body that is $\{a\}$ the $\{S_m\}$ of the $\{c\}$. Show the whole person with natural pose, contact, perspective, and shadows. Do not alter the $\{c\}$'s geometry, position or components.}
\end{quote}

However, not all affordances are naturally expressed through a human body.
For affordances that are better demonstrated by another object (\eg, a `table' affords `support'), we use an object-object interaction branch.
In this case, we first ask the LLM to select a concrete, reconstructable interactor object $u_{c,a}$ for the pair $(c,a)$:
\begin{quote}
\small
\textit{Target object class: $\{c\}$. Target affordance: $\{a\}$. Choose one physical, discrete solid object that would best demonstrate this affordance when inserted into an image. The object must be reconstructable in 3D (clear boundaries, stable shape), and must NOT be a liquid, gas, powder, amorphous material, or effect (e.g., no water, smoke, fire, sand, mud, foam, light, shadow). Return only one object noun phrase in lowercase (1 word), no explanation.}
\end{quote}
After obtaining $u_{c,a}$, we synthesize the object-object interaction using:
\begin{quote}
\small
\textit{Edit this image to add a realistic $\{u_{c,a}\}$ that is $\{a\}$ the $\{S_m\}$ of the $\{c\}$. The inserted $\{u_{c,a}\}$ must have natural contact, perspective, and shadows. Do not alter the $\{c\}$'s geometry, position or components.}
\end{quote}
Thus, each synthesized image $\mI_m$ contains either a human-object interaction or an object-object interaction, depending on the affordance category.
In the following sections, we use human-object interaction by default for illustration.

\myparagraph{Depth Alignment.}
After we obtain the object point cloud $\gO = \{\vo_i\}_{i=1}^{N_o}$ and a human point cloud $\gH_m = \{\vh_j\}_{j=1}^{N_h}$, we use DepthAnything3~\cite{lin2025depth} to estimate a depth map $\mD$ for the HOI image $\mI_m$.
To avoid occlusion-induced depth errors, we only use reconstructed points whose projections fall inside the corresponding visible 2D masks. 
Specifically, object points are fitted using $\{\vo_i \in \gO \mid \pi(\vo_i) \in \mM^o\}$, and human points are fitted using $\{\vh_j \in \gH_m \mid \pi(\vh_j) \in \mM_m^h\}$, where $\mM^o$ and $\mM_m^h$ denote the visible object and human masks in the HOI image generated by SAM3~\citep{carion2025sam}.
Therefore, occluded object points that are not visible in the image are excluded from the affine depth alignment.
Let $\pi(\vv)$ denote the projection of a 3D point $\vv$ onto the image plane.
For the object and human point clouds, we estimate separate affine mappings from reconstructed depth to monocular depth:
\bea
(a^o, b^o) = 
\mathop{\arg\min}\limits_{a, b}
\sum_{\vo_i \in \gO:\,\pi(\vo_i)\in\mM^o}
\left\|
a z^o_i + b - \mD(\pi(\vo_i))
\right\|_2^2 \\
(a^h, b^h) = \mathop{\arg\min}\limits_{a, b} \sum_{\vh_j \in \gH_m:\,\pi(\vh_j)\in\mM_m^h} \left\| a z^h_j + b - \mD(\pi(\vh_j)) \right\|_2^2,
\eea
where each object point is $\vo_i = (x_i^o, y_i^o, z_i^o)$ and each human point is $\vh_j = (x_j^h, y_j^h, z_j^h)$.
We keep the object geometry fixed and transform the human depth into the object depth scale.
For each human point $\vh_j=(x^h_j,y^h_j,z^h_j)$, its aligned depth is:
\bea
\tilde{z}^h_j = \frac{a^h z^h_j + b^h - b^o}{a^o}.
\eea
To preserve the 2D projection, we move each human point only along its original camera ray:
\bea \tilde{\vh}_j = \left( \frac{x^h_j}{z^h_j}\tilde{z}^h_j, \frac{y^h_j}{z^h_j}\tilde{z}^h_j, \tilde{z}^h_j \right).
\eea
This gives the aligned human point cloud $\tilde{\gH_m}=\{\tilde{\vh}_j\}_{j=1}^{N_h}$, which is placed in the same 3D coordinate system as the object point cloud $\gO$.

\myparagraph{Region-guided keypoint selection.}
Given the aligned object-human point clouds $(\gO,\tilde{\gH})$, we infer the interaction region on the object by measuring its proximity to the human.
For each object point $\vo_i \in \gO$, we compute its nearest-neighbor distance to the aligned human point cloud:
\bea
d_i = \min_{\tilde{\vh}_j \in \tilde{\gH}} \left\| \vo_i - \tilde{\vh}_j \right\|_2.
\eea
We define the candidate affordance region as the set of object points whose distances are below a threshold:
\bea
\gR_m = \left\{ \vo_i \in \gO \mid d_i \leq \tau_d \right\},
\eea
where $\tau_d$ controls the spatial extent of the interaction region.
In practice, we set $\tau_d$ to the $30$th percentile of all point-wise distances $\{d_i\}_{i=1}^{N_o}$.
Finally, we sample $P$ keypoints from $\gR_m$ and follow the label propagation rule of prior 3D affordance annotations~\cite{deng20213d} to obtain a soft point-level affordance annotation $\mY_m$ for the $m$-th interaction pattern.
Aggregating the annotations from all synthesized HOI images $\mI_m$ yields a multi-mask annotation set $\{\mY_m\}_{m=1}^{M}$.

\begin{table}[t]
\centering
\caption{\textbf{Statistics of \DataName{}.}
Train and Val denote the number of shapes in each split. Ann. denotes the total number of point-level affordance masks, including the original 3DAffordanceNet annotations and our curated annotations.}
\label{tab:dataset_pair_stats}
\scriptsize
\setlength{\tabcolsep}{3pt}
\resizebox{\linewidth}{!}{
    \begin{tabular}{llrrr@{\hspace{0.8em}}llrrr}
    \specialrule{.15em}{.05em}{.05em}
    \multicolumn{5}{c}{Object-affordance pairs} & \multicolumn{5}{c}{Object-affordance pairs} \\
    \cmidrule(lr){1-5} \cmidrule(lr){6-10}
    Object & Aff. & Train & Val & Ann. & Object & Aff. & Train & Val & Ann. \\
    \midrule
    Chair & move & 980 & 599 & 5,546 & Hat & grasp & 149 & 20 & 495 \\
    Chair & support & 979 & 597 & 5,337 & Microwave & contain & 125 & 17 & 471 \\
    Table & move & 839 & 636 & 4,916 & TrashCan & openable & 138 & 18 & 471 \\
    Chair & sittable & 988 & 594 & 4,298 & Refrigerator & contain & 112 & 17 & 446 \\
    Table & support & 745 & 592 & 4,186 & Earphone & listen & 157 & 22 & 444 \\
    StorageFurniture & contain & 853 & 127 & 3,116 & Bowl & wrap\_grasp & 129 & 17 & 436 \\
    StorageFurniture & openable & 796 & 118 & 2,379 & Door & pushable & 128 & 20 & 417 \\
    Vase & contain & 561 & 79 & 2,311 & Hat & wear & 137 & 18 & 398 \\
    Vase & wrap\_grasp & 590 & 84 & 2,156 & Door & openable & 123 & 19 & 377 \\
    Vase & pourable & 560 & 80 & 1,717 & Bed & layable & 107 & 14 & 374 \\
    Faucet & grasp & 394 & 61 & 1,378 & Bowl & pourable & 124 & 17 & 366 \\
    Faucet & openable & 379 & 60 & 1,144 & Bag & contain & 87 & 12 & 356 \\
    Laptop & press & 277 & 41 & 897 & Mug & pourable & 119 & 18 & 352 \\
    Bottle & contain & 238 & 34 & 860 & Microwave & openable & 115 & 18 & 350 \\
    TrashCan & contain & 204 & 27 & 842 & Bed & support & 98 & 14 & 341 \\
    Bottle & wrap\_grasp & 271 & 37 & 807 & Dishwasher & contain & 99 & 14 & 336 \\
    Bottle & grasp & 210 & 33 & 790 & Door & pull & 105 & 15 & 327 \\
    Bottle & pourable & 248 & 34 & 673 & Refrigerator & openable & 116 & 15 & 317 \\
    Bottle & openable & 240 & 35 & 647 & Bed & sittable & 104 & 13 & 312 \\
    TrashCan & pourable & 204 & 27 & 599 & Keyboard & press & 108 & 15 & 285 \\
    Knife & stab & 187 & 25 & 545 & Dishwasher & openable & 109 & 15 & 274 \\
    Knife & grasp & 176 & 22 & 540 & Bag & lift & 76 & 9 & 251 \\
    Mug & wrap\_grasp & 130 & 19 & 532 & Bag & grasp & 84 & 10 & 232 \\
    Bowl & contain & 126 & 17 & 526 & Bag & openable & 71 & 10 & 220 \\
    Knife & cut & 181 & 22 & 517 & Scissors & grasp & 47 & 6 & 150 \\
    Mug & contain & 123 & 18 & 517 & Scissors & stab & 41 & 6 & 142 \\
    Earphone & grasp & 153 & 22 & 512 & Scissors & cut & 39 & 5 & 131 \\
    Mug & grasp & 127 & 18 & 498 &  &  &  &  &  \\
    \specialrule{.15em}{.05em}{.05em}
    \end{tabular}
}
\label{tab:supp_dataset_stats}
\end{table}

\section{\DataName{} details}
\label{sec:dataset_details}
\myparagraph{Dataset construction.}
We construct \DataName{} starting from the original 3DAffordanceNet~\citep{deng20213d} annotations.
Since our goal is to generate 3D affordance annotations from physical human-object interactions, we remove object-affordance pairs whose affordance does not correspond to a direct physical interaction.
Specifically, we discard the non-interactive affordance \emph{display} for three object-affordance pairs: Display-display, Clock-display, and Laptop-display.
In addition, to reduce dataset imbalance, we cap the number of objects to at most 1,000 for each object category.

\DataName{} contains 14,606 training samples and 4,452 validation samples, covering 55 object-affordance pairs, 21 object categories, and 17 affordance labels.
In addition to the original 3DAffordanceNet annotations, our pipeline generates 29,274 additional masks for training and 9,525 additional masks for validation.
We summarize the object-affordance pair statistics in~\tabref{tab:supp_dataset_stats}.

\myparagraph{Evaluation details.}
The original 3DAffordanceNet annotations are created by manually selecting sparse keypoints on each 3D shape and propagating them into dense soft affordance maps~\citep{deng20213d}.
However, the raw human-selected keypoints and the exact propagation parameters are not fully released.
Therefore, we recover approximate human keypoints from the provided affordance maps by selecting high-response local maxima on the 3D affordance masks.
When comparing our generated annotations with the recovered ground-truth keypoints, we account for the multi-mask nature of our data by evaluating the best-matching generated mode.
Specifically, we compute the keypoint distance for each generated mode and report the minimum distance to the recovered ground truth.

\myparagraph{Additional quantitative results.}
To further quantify the spatial agreement between our generated keypoints and the recovered 3DAffordanceNet keypoints, we report the percentage of samples for which at least half of the generated keypoints fall within a normalized distance threshold from the recovered ground-truth keypoints.
Using normalized distance thresholds of 0.1, 0.2, and 0.3, the corresponding percentages are 31.1\%, 67.6\%, and 87.0\%, respectively.
These results indicate that most generated annotations are spatially close to the recovered human annotations, while still allowing valid variation across different interaction modes.

\section{Baseline implementation details}
\label{sec:baseline_details}
\subsection{Deterministic baselines.}
We compare against PointRefer~\citep{li2024laso}, 3D-ADLLM~\citep{chu20253d}, and GEAL~\citep{lu2025} as representative deterministic 3D affordance grounding baselines.
For the standard single-mask benchmark, we directly evaluate their predicted masks using single-mask evaluation metrics.
For the multi-mask benchmark, we treat each deterministic prediction as a singleton prediction set.

In the multi-mask setting, we train PointRefer and GEAL on our \DataName{} following their official implementations and training settings.
In the single-mask setting, we directly evaluate the official GEAL checkpoint, which is trained with LASO-seen~\citep{li2024laso} data.
For PointRefer, since official checkpoints are not provided, we retrain the model using the official training scripts.
For 3D-ADLLM, since the released training scripts are not directly reproducible, we use the official checkpoint and evaluate it in both the single-mask and multi-mask settings.
Note, the official 3D-ADLLM checkpoint is trained on IRAS~\citep{chu20253d}, which differs from the LASO dataset~\citep{li2024laso}.

\subsection{Generative baselines.}

\myparagraph{PointRefer-CVAE.}
We implement PointRefer-CVAE by replacing the deterministic PointRefer mask decoder with a conditional variational decoder.

Given the PointRefer backbone features, we first construct a global context feature by pooling the point features and averaging the text-token features, followed by an MLP projection.
The prior network predicts a diagonal Gaussian distribution $p_{\theta}(\vz\mid \mP,\mA)$ from this context feature.
During training, the posterior network additionally takes a ground-truth affordance mask as input and predicts $q_{\phi}(\vz\mid \mP,\mA,\mY^*)$.
A latent code is sampled using the reparameterization trick and decoded into a point-level affordance mask.
The decoder predicts a mask embedding from the context and latent code, and also predicts a latent-conditioned scale and bias to modulate the point features.
The final mask is obtained by the inner product between the mask embedding and the modulated point features, followed by a sigmoid activation.
PointRefer-CVAE is trained with the objective:
\bea
\mathcal{L}_{\text{CVAE}}=\lambda_{\text{bce}}\ell_{\text{bce}}+\lambda_{\text{dice}}\ell_{\text{dice}}+\lambda_{\text{kl}}D_{\text{KL}}(q_{\phi}(\vz\mid\mP,\mA,\mY^*)\|p_{\theta}(\vz\mid\mP,\mA)).
\eea
We compute the reconstruction loss using the posterior samples during training.
At inference time, we sample latent codes from the learned prior and decode multiple masks for each input.
We use a latent dimension of $64$, hidden dimension of $512$, and generate $4$ samples per input for both training and evaluation.
The BCE, Dice, and KL weights are set to $1.0$, $1.0$, and $0.1$, respectively.

\myparagraph{PointRefer-LDM.}
We implement PointRefer-LDM as a latent diffusion baseline built on the same PointRefer backbone.
The model contains a mask autoencoder and a latent diffusion model.

The mask encoder maps each ground-truth affordance mask into a latent representation.
The mask decoder maps a latent code back to a point-level affordance mask by predicting a mask embedding and latent-conditioned point-feature modulation.
We first train this mask autoencoder using BCE and Dice reconstruction losses.
The autoencoder objective is
\bea 
\mathcal{L}_{\text{AE}}=\lambda_{\text{bce}}\ell_{\text{bce}}+\lambda_{\text{dice}}\ell_{\text{dice}}.
\eea
In our experiments, we set $\lambda_{\text{bce}}=1.0$, $\lambda_{\text{dice}}=1.0$.

After training the autoencoder, we train a diffusion model in the learned latent space.
The diffusion model is conditioned on the same global context feature from the PointRefer backbone.
At each training iteration, Gaussian noise is added to the clean mask latent according to a sampled timestep, and the denoising network predicts the injected noise.
We use the latent diffusion objective:
\bea
\mathcal{L}_{\text{LDM}}=\|\epsilon-\hat{\epsilon}\|_2^2.
\eea
During inference, we sample Gaussian noise and denoise it for $50$ steps to obtain latent samples, which are then decoded into affordance masks.
We use a latent dimension of $64$, hidden dimension of $512$, sinusoidal timestep embedding dimension of $512$, and generate $4$ samples per input.
The diffusion noise schedule uses linearly spaced $\beta_t$ values from $10^{-4}$ to $2\times10^{-2}$.
The first-stage autoencoder is frozen when training the diffusion model.

During inference, each generative baseline produces $4$ affordance masks per input, which matches the maximum number of valid annotation modes for an instance in \DataName{}.

\section{Evaluation metrics}
\label{sec:supp_eval_metrics}
\subsection{Single-mask evaluation metrics.}
Given a point cloud with $N$ points, we denote the predicted affordance mask as $\mY \in [0,1]^N$ and the ground-truth affordance mask as $\mY^* \in [0,1]^N$.
For single-mask evaluation, each test sample contains one ground-truth mask.
When a model predicts a set of masks $\{\mY_s\}_{s=1}^{S}$, we select the prediction with the lowest MAE to the ground-truth mask $\mY^*$ and report all single-mask metrics on this selected mask.

\myparagraph{AUC.}
We compute the Area Under the ROC Curve (AUC) by first binarizing the ground-truth affordance mask:
\bea
\bar{\mY}^{*}_i = \mathbbm{1}[\mY_i^* \geq 0.5],
\eea
where $\mathbbm{1}[\cdot]$ is the indicator function.
AUC measures whether the predicted affordance scores $\mY$ rank positive points higher than negative points:
\bea
\operatorname{AUC}(\mY, \mY^*) =
\operatorname{ROC\text{-}AUC}(\mY, \bar{\mY}^{*}).
\eea
A higher AUC indicates better point-wise affordance discrimination.

\myparagraph{mIoU.}
We compute mean Intersection-over-Union (mIoU) by averaging IoU over multiple prediction thresholds.
For each threshold $\tau \in \mathcal{T}$, where $\mathcal{T}$ contains 20 uniformly spaced thresholds in $[0,1]$, we binarize the prediction as:
\bea
\bar{\mY}^{\tau}_i = \mathbbm{1}[\mY_i \geq \tau].
\eea
The IoU at threshold $\tau$ is:
\bea
\mathrm{IoU}_{\tau}(\mY, \mY^*) =
\frac{
\sum_i \bar{\mY}^{\tau}_i \bar{\mY}^{*}_i
}{
\sum_i \mathbbm{1}[\bar{\mY}^{\tau}_i + \bar{\mY}^{*}_i > 0]
},
\eea
where we use the same binarized ground-truth mask $\bar{\mY}^{*}_i=\mathbbm{1}[\mY_i^*\geq0.5]$.
The final mIoU is:
\bea
\operatorname{mIoU}(\mY, \mY^*)=
\frac{1}{|\mathcal{T}|}
\sum_{\tau \in \mathcal{T}}
\operatorname{IoU}_{\tau}(\mY, \mY^*).
\eea
A higher mIoU indicates better spatial overlap with the ground-truth affordance region.

\myparagraph{SIM.}
Similarity (SIM) measures the overlap between normalized affordance distributions.
We first normalize the prediction and ground truth:
\bea
\tilde{\mY}_i =
\frac{\mY_i}{\sum_{i'} \mY_{i'}},
\quad
\tilde{\mY}^{*}_i =
\frac{\mY_i^*}{\sum_{i'} \mY_{i'}^*}.
\eea
SIM is computed as:
\bea
\operatorname{SIM}(\mY, \mY^*)=
\sum_i \min(\tilde{\mY}_i, \tilde{\mY}^{*}_i).
\eea
A higher SIM indicates stronger distributional agreement between the predicted and ground-truth affordance maps.

\myparagraph{MAE.}
Mean Absolute Error (MAE) measures the average point-wise prediction error:
\bea
\operatorname{MAE}(\mY, \mY^*)=
\frac{1}{N}
\sum_{i=1}^{N}
\left|\mY_i - \mY_i^*
\right|.
\eea
A lower MAE indicates stronger point-wise agreement.

\subsection{Multi-mask evaluation metrics.}
For multi-mask evaluation, each test sample contains a set of plausible ground-truth affordance masks $\mathcal{Y}=\{\mY_k^*\}
_{k=1}^{K}$.
The model predicts a set of affordance masks $\{\mY_s\}_{s=1}^{S}$.
Since multiple affordance regions may be valid for the same object and affordance query, we use Generalized Energy Distance (GED)~\citep{ribeiro2025flow} to evaluate whether the predicted mask set matches the ground-truth mask distribution.
GED is defined as:
\bea
D^2_{\text{GED}}(p_{\text{data}}, p_{\text{model}})
&=&
2\sE_{\mY^* \sim p_{\text{data}}, \mY \sim p_{\text{model}}}
\left[d(\mY^*,\mY)\right] -
\sE_{\mY,\mY' \sim p_{\text{model}}}
\left[d(\mY,\mY')\right]
\nonumber\\
&&
- \sE_{\mY^*,{\mY^*}' \sim p_{\text{data}}}
\left[d(\mY^{*},{\mY^*}')\right].
\eea
where $d(\cdot,\cdot)$ is a task-specific distance between two point-wise affordance masks.
In practice, we estimate GED using the predicted mask set $\{\mY_s\}_{s=1}^{S}$ and the ground-truth mask set $\mathcal{Y}=\{\mY_k^*\}_{k=1}^{K}$:
\bea
D^2_{\text{GED}} =
\frac{2}{KS}
\sum_{k=1}^{K}
\sum_{s=1}^{S}
d(\mY_k^*, \mY_s) -
\frac{1}{S^2}
\sum_{s=1}^{S}
\sum_{s'=1}^{S}
d(\mY_s, \mY_{s'}) -
\frac{1}{K^2}
\sum_{k=1}^{K}
\sum_{k'=1}^{K}
d(\mY_k^*, \mY_{k'}^*).
\eea
The first term measures the discrepancy between predicted and ground-truth masks.
The second and third terms account for the diversity within the predicted masks and ground-truth masks respectively.
A lower GED indicates that the predicted mask distribution better matches the set of plausible ground-truth affordance regions.

We instantiate $d(\cdot,\cdot)$ using distances derived from the single-mask metrics.
Since GED requires the underlying distance to have zero self-distance, \ie, $d(\mY,\mY)=0$, we use MAE-, SIM-, and AUC-based distances.
For MAE, we directly use:
\bea
d_{\operatorname{MAE}}(\mY_s, \mY_k^*)=
\operatorname{MAE}(\mY_s, \mY_k^*).
\eea
For SIM, we use:
\bea
d_{\operatorname{SIM}}(\mY_s, \mY_k^*)=
1 - \operatorname{SIM}(\mY_s, \mY_k^*).
\eea
For AUC, the score depends on which mask is treated as the prediction and which mask is treated as the reference.
We therefore use symmetric distances:
\bea
d_{\operatorname{AUC}}(\mY_s, \mY_k^*)=
1 -
\frac{
\operatorname{AUC}(\mY_s, \mY_k^*)
+
\operatorname{AUC}(\mY_k^*, \mY_s)
}{2}.
\eea
Although mIoU is used in single-mask evaluation, we do not adapt it to GED, as it violates the zero self-distance requirement.
This gives three multi-mask metrics: GED-MAE, GED-SIM, and GED-AUC.
For all GED variants, \emph{lower} values indicate better multi-mask prediction quality.

\begin{figure}[t]
\centering
\includegraphics[width=0.8\linewidth]{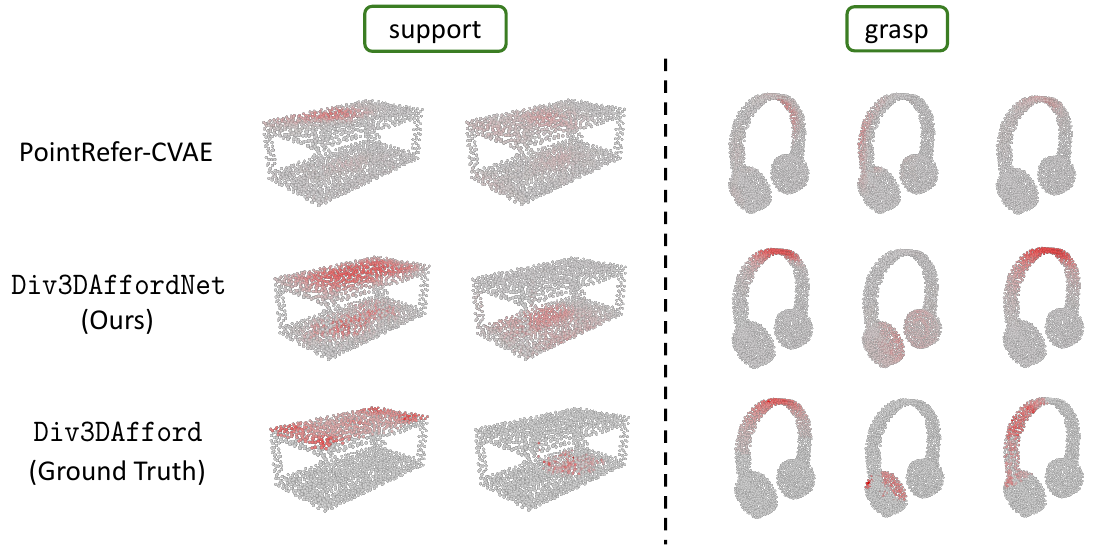}
\caption{\textbf{Additional qualitative comparison on diverse 3D affordance grounding.}
We compare \ModelName{} with the PointRefer-CVAE baseline on the \DataName{} validation set.
For each object-affordance query, we visualize representative predictions for different output modes.
Red indicates the predicted affordance region.
}
\label{fig:supp_qual_multi_results}
\end{figure}

\begin{figure}[t]
\centering
\includegraphics[width=0.5\linewidth]{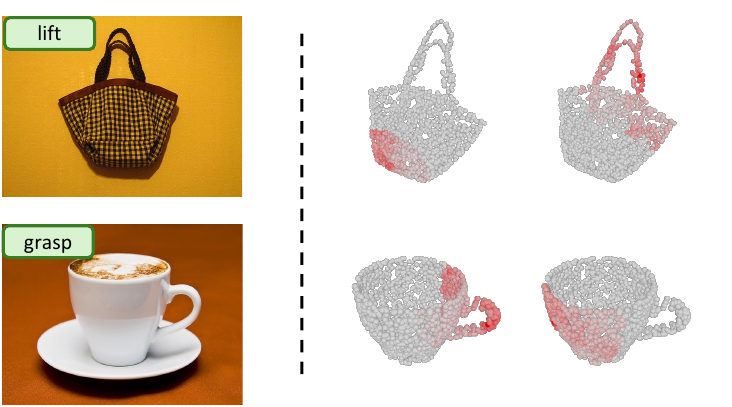}
\caption{\textbf{Generation pipeline on 2D images.}
We show examples on an ImageNet~\citep{deng2009imagenet} bag and an AGD20K~\citep{luo2022learning} cup.
Given a 2D image and an affordance query, we reconstruct the 3D object geometry and generate multiple plausible 3D affordance annotations, capturing diverse interaction regions.
}
\label{fig:qual_other_data_gen}
\end{figure}

\section{Additional qualitative results}
\label{sec:add_qual_results}
\myparagraph{\ModelName{}.}
We visualize additional qualitative results in~\figref{fig:supp_qual_multi_results}.
Across different object categories and affordance queries, \ModelName{} predicts multiple spatially distinct affordance regions that align well with the multi-mask ground-truth annotations.
Compared with PointRefer-CVAE, our predictions are more localized and less noisy, indicating that the query-based set prediction design better captures discrete interaction modes.

\myparagraph{Generation Pipeline.}
We further apply our proposed pipeline to 2D images from ImageNet~\citep{deng2009imagenet} and AGD20K~\citep{luo2022learning}, with examples shown in~\figref{fig:qual_other_data_gen}.
The results show that our approach can effectively construct multi-mask 3D affordance annotations directly from diverse 2D images.

\section{Additional ablation studies}
\label{sec:add_ablation_studies}
We conduct an additional ablation study on the positional embedding used in \ModelName{}.
Specifically, we replace our Fourier positional embedding with the sinusoidal positional embedding used in 2D DETR~\cite{carion2020end}.
This variant achieves a GED-MAE of 0.096, which is the same as our default Fourier positional embedding.
This result suggests that the choice of positional embedding is not critical for our diverse 3D affordance grounding task.
Therefore, we keep the Fourier positional embedding, following 3DETR~\cite{misra2021end} for 3D point cloud inputs.

\begin{table}[t]
\centering
\caption{
\textbf{Computational cost comparison.}
For PointRefer, the raw profiling result corresponds to three full runs; we divide its measured time and FLOPs by three to report the cost of a single forward run.
}
\begin{tabular}{lcc}
\specialrule{.15em}{.05em}{.05em}
\multirow{2}{*}{Method} & Inference time & FLOPs \\
                        & (ms / sample) & (GFLOPs / sample) \\
\midrule
PointRefer & 8.08 & 24.01 \\
PointRefer-CVAE & 19.51 & 24.21 \\
\colorours \ModelName{} & 20.57 & 29.21 \\
\specialrule{.15em}{.05em}{.05em}
\end{tabular}
\label{tab:supp_computational_cost}
\end{table}

\section{Computational cost}
\label{sec:compute}
We report the inference cost of \ModelName{} against both single-mask and multi-mask baselines in~\tabref{tab:supp_computational_cost}.
We compare the per-sample inference time and FLOPs on the same NVIDIA L40S GPU.
For PointRefer, the official PointRefer implementation averages predictions over three repeated forward runs.
We divide the measured latency and FLOPs by three to report the cost of a single forward pass.
Our \ModelName{} requires 20.57 ms and 29.21 GFLOPs per sample.
This is comparable to the PointRefer-CVAE baseline, which requires 19.51 ms per sample and 24.21 GFLOPs per sample.
Although PointRefer has a lower single-run latency of 8.08 ms per sample, it is a deterministic baseline and does not explicitly decode a set of diverse affordance masks in one forward pass.
In contrast, \ModelName{} predicts multiple plausible affordance regions jointly through its set-prediction decoder, while maintaining a moderate computational overhead.

\section{Limitations}
\label{sec:limitations}

\myparagraph{Annotation pipeline.}
Although our annotation generation pipeline enables scalable multi-mask 3D affordance annotation, it inherits limitations from the underlying 2D generative and 3D reconstruction models.
Image editing may synthesize implausible interactions or alter fine-grained object geometry, while reconstruction may introduce incomplete geometry, shape distortion, or human-object misalignment.
Although we mitigate these issues through alignment and manual filtering, noisy pseudo annotations may still remain.

\myparagraph{\DataName{} coverage.}
\DataName{} is constrained by the affordance labels and object categories inherited from 3DAffordanceNet, and therefore does not cover the full diversity of real-world objects, interactions, or task-specific affordance semantics.
Moreover, the generated annotations are not exhaustive; additional valid affordance modes may exist beyond those captured by our synthesized interactions.

\end{document}